\documentclass[10pt,twocolumn,letterpaper]{article}

\usepackage{cvpr}
\usepackage{times}
\usepackage{epsfig}
\usepackage{graphicx}
\usepackage{amsmath}
\usepackage{amssymb}
\usepackage{booktabs} % To thicken table lines

\usepackage[ruled,linesnumbered]{algorithm2e}

\usepackage[pagebackref=true,breaklinks=true,letterpaper=true,colorlinks,bookmarks=false]{hyperref}

 \cvprfinalcopy % *** Uncomment this line for the final submission

\def\cvprPaperID{419} % *** Enter the CVPR Paper ID here

\ifcvprfinal\fi
\begin{document}

%%%%%%%%% TITLE
\title{Learning Chained Deep Features and Classifiers for Cascade in Object Detection}

\author{Wanli Ouyang,  Ku Wang, Xin Zhu, Xiaogang Wang \\
The Chinese University of Hong Kong\\
{\tt\small \{wlouyang, kwang, xzhu, xgwang\}@ee.cuhk.edu.hk}
% For a paper whose authors are all at the same institution,
% omit the following lines up until the closing ``}''.
% Additional authors and addresses can be added with ``\and'',
% just like the second author.
% To save space, use either the email address or home page, not both
}

\maketitle
%\thispagestyle{empty}

%%%%%%%%% ABSTRACT
\begin{abstract}
Cascade is a widely used approach that rejects obvious negative samples at early stages for learning better classifier and faster inference.
This paper presents chained cascade network (CC-Net). In this CC-Net, the cascaded classifier at a stage is aided by the classification scores in previous stages. Feature chaining is further proposed so that the feature learning for the current cascade stage uses the features in previous stages as the prior information. The chained ConvNet features and classifiers of multiple stages are jointly learned in an end-to-end network. In this way, features and classifiers at latter stages handle more difficult samples with the help of features and classifiers in previous stages.  It yields consistent boost in detection performance on benchmarks like PASCAL VOC 2007 and ImageNet.
Combined with better region proposal, CC-Net leads to state-of-the-art result of 81.1\% mAP on PASCAL VOC 2007.
\end{abstract}

%%%%%%%%% BODY TEXT
\section{Introduction}
Object detection is a fundamental computer vision task. 
It differs from image classification in that the number of background samples (image regions not belonging to any object class of interest) is much larger than the number of object samples. This leads to the undesirable imbalance in the number of samples for different classes during training.

In order to handle the imbalance problem from the background samples, bootstrapping, cascade, and hard negative mining have been developed \cite{sung1996learning,Dalal:HOG,Viola:HaarConf}. In these approaches, classifier learning is divided into multiple stages. In each stage, only a subset of background samples are used for training. The classifiers at earlier stages handle easier samples while the classifiers at latter stages handle more difficult samples. Bootstrapping and hard negative mining aims at learning more accurate classifier.  In comparison, cascade improves both accuracy and speed of the detection process by rejecting easy background samples at both training and testing time. 

In bootstrapping, cascade, and hard negative mining, however, most of the information obtained from evaluating a given stage is discarded in the next stage. The decision in the current stage does not take the detection confidence in the previous stage into account. Take the image in Fig \ref{fig:overview} as an example, since the region proposal is not accurate, it might find a region-of-interest (RoI) that looks like a horse or antelope. The classifier at the first stage uses the visual cue decide that the RoI may be an antelope or a horse. Then the classifier at the second stage uses the visual cue with more contextual region to decide that it may be a cattle or an antelope. The confidence for antelope in the first stage is not  passed to the second stage in these  approaches. The classifier in each stage is imperfect. A sample might be wrongly treated as background at the second stage although the classifier at the first stage is confident in seeing an object.

\begin{figure}
\begin{center}
\centerline{\epsfig{figure=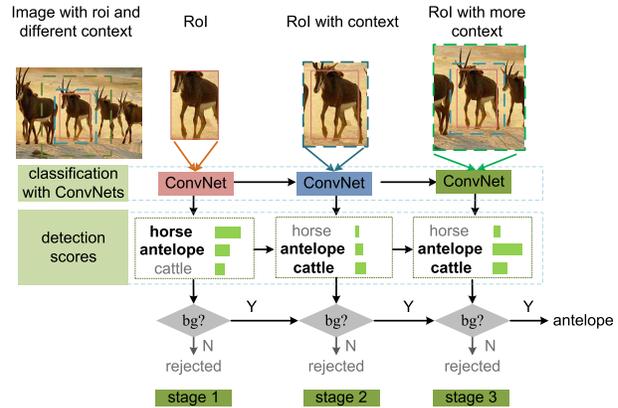,width=8cm}}
\end{center}
\vspace{-10pt}
   \caption{Motivation of the chained cascade ConvNet in chaining both classification scores and features. In the image, rectangle  in solid line denotes RoI and rectangles in dashed lines denote contextual region. Scores and features at previous stages are useful for latter stages. A chaining structure is used for learning classifier and feature based on the prior knowledge of the classifiers and features in the previous stages. Best viewed in color.}
\label{fig:overview}
\end{figure}

The design of chaining classification scores among cascade stages is called soft cascade \cite{bourdev2005robust} and boosting chain in \cite{xiao2003boosting}. In these approaches, the detection scores in the previous stages are added to the detection scores in the current stage. With this design, the confidence for the antelope at the first stage is used for the classifier in the second stage for the example in Fig \ref{fig:overview}. Chaining detection scores improves the robustness of the cascaded classifiers.
It is found to be effective in face detection \cite{bourdev2005robust, xiao2003boosting} and generic object detection \cite{Pedersoli:C2FObj2, Felzenszwalb:cascade10} for hand crafted features. 

Deep ConvNets have led to significant gains in accuracy for object detection \cite{girshick2014rich, girshick2015fast}. Granted with the ability of ConvNets in learning multiple components end-to-end \cite{wan2015end}, we aim at learning the cascade of chained classifiers and chained features with an end-to-end network.

The motivation of this paper is two folds:

First, joint learning of chained classifiers in multiple cascade stages for generic object detection by single ConvNet. Through communication with detection scores and a joint loss function, the classifiers in multiple stages can cooperate with each other in both feature learning and classifier learning. For example, when the first classifier finds that the object should only be a mammal, then the features and classifier at the second stage can focus more on distinguishing specific class of mammal like horse, antelope and cattle. 

Second, the feature learning of the current stage should be based on the prior knowledge of the features in the previous stage. For example, when the mammal body and head are found at the first stage and the horns are found at the second stage, the feature learning at the third stage should focus more on features that help to distinguish mammals with horns. Therefore, we design feature chaining to utilize the features in previous stages.

Based on the observations above, we design a chained cascade network (CC-Net) for object detection. The contribution of this design is as follows:
\begin{enumerate}
\item The network jointly learns the cascade of multiple chained classifiers. Cascade facilitates the learning of more powerful features in latter stages by rejecting background samples at early stages.
 By chaining classifiers, the classification results in previous stages serve as prior information for the classification at the current stage.
\item We design feature chaining so that the feature learning at the current stage utilizes the prior knowledge of features in previous stages. In this way, the feature learning at the current stage focuses more on the visual cues that are complementary to previous stages.
\item Classifier chaining and feature chaining are jointly learned using a single ConvNet.
\end{enumerate}
Our design is shown to be effective even when only 300 boxes are retained for each image after using state-of-the-art region proposal approaches.
Experimental results on ImageNet and Pascal VOC 2007 show improvement in detection accuracy by 5.1\% and 3.5\% in mean average precision (mAP) respectively.

\section{Related work}
\emph{Cascade, bootstrapping and hard example mining}. Bootstrapping was introduced in the work of Sung
and Poggio \cite{sung1996learning} in the mid-1990’s for training face detection models. With the success of HOG+SVM+bootstrapping based methods \cite{Dalal:HOG, LatSVMObj}, bootstrapping was frequently
used when training SVMs for object detection \cite{girshick2014rich}. Felzenszwalb \etal \cite{LatSVMObj} proved that a form of bootstrapping
for SVMs converges to the global optimal solution defined on the entire dataset. Their algorithm was often
referred to as hard negative mining.
Cascade has appeared in various forms dating back to the 1970s, as was pointed out by Schneiderman \cite{schneiderman2004feature}.
It has been widely used in object detection \cite{Pedersoli:C2FObj2, Felzenszwalb:cascade10,bourdev2005robust,dollar2014fast,li2004floatboost}. Cascade can be applied for SVM \cite{Pedersoli:C2FObj2, Felzenszwalb:cascade10}, boosted classifiers \cite{dollar2014fast,li2004floatboost,xiao2003boosting}, and ConvNets \cite{yang2016craft}. Although not explicitly stated, the use of region proposal followed by RCNN or fast RCNN can also be considered as cascade, in which region proposal rejects background samples and fast RCNN provides classification scores. As a summary,  classifiers are learned stage by stage for the approaches mentioned above. The communication among the classifiers are based on samples rejection or detection scores.

 Recently, Shrivastava \etal introduced online mining of hard positive and negative examples for ConvNet-based detector \cite{shrivastava2016training} . In this approach, the learning of classifiers at the first stage in rejecting easy samples and the second stage in obtaining detection scores were merged into a single learning stage.  
Qin \etal proposed a joint training of cascaded classifier for face detection using ConvNet \cite{qin2016joint}. Our approach is different from them in two aspects. First, the effectiveness of joint learning of cascade is found to be effective for face detection in \cite{qin2016joint} but unknown for generic object detection. Second, the chaining of scores and features in multiple stages is not built up in \cite{qin2016joint, shrivastava2016training} but is the main objective in our approach.

Deeper ConvNets were found to be effective for image classification and object detection \cite{Krizhevsky:ImageNetCNN,sermanet2013overfeat,simonyan2014very,szegedy2015going,he2016deep}. On other hand, wide residual network \cite{zagoruyko2016wide}, inception modules \cite{szegedy2015going, chollet2016deep}, multi-region features \cite{gidaris2015object, zeng2016gated,bell2015inside}  showed that increasing the width of the ConvNets in an effective way led to improvement the image classification accuracy. Our work is complementary to the works above that learn better features. We can use these features to obtain diverse features for cascade in different stages. In our work, features of the same depth are divided into different cascade stages and  communicate by feature chaining. This design, which is not investigated in previous works, improves the ability of features in handling more difficult examples in latter cascade stages.

\begin{figure*}
\begin{center}
\centerline{\epsfig{figure=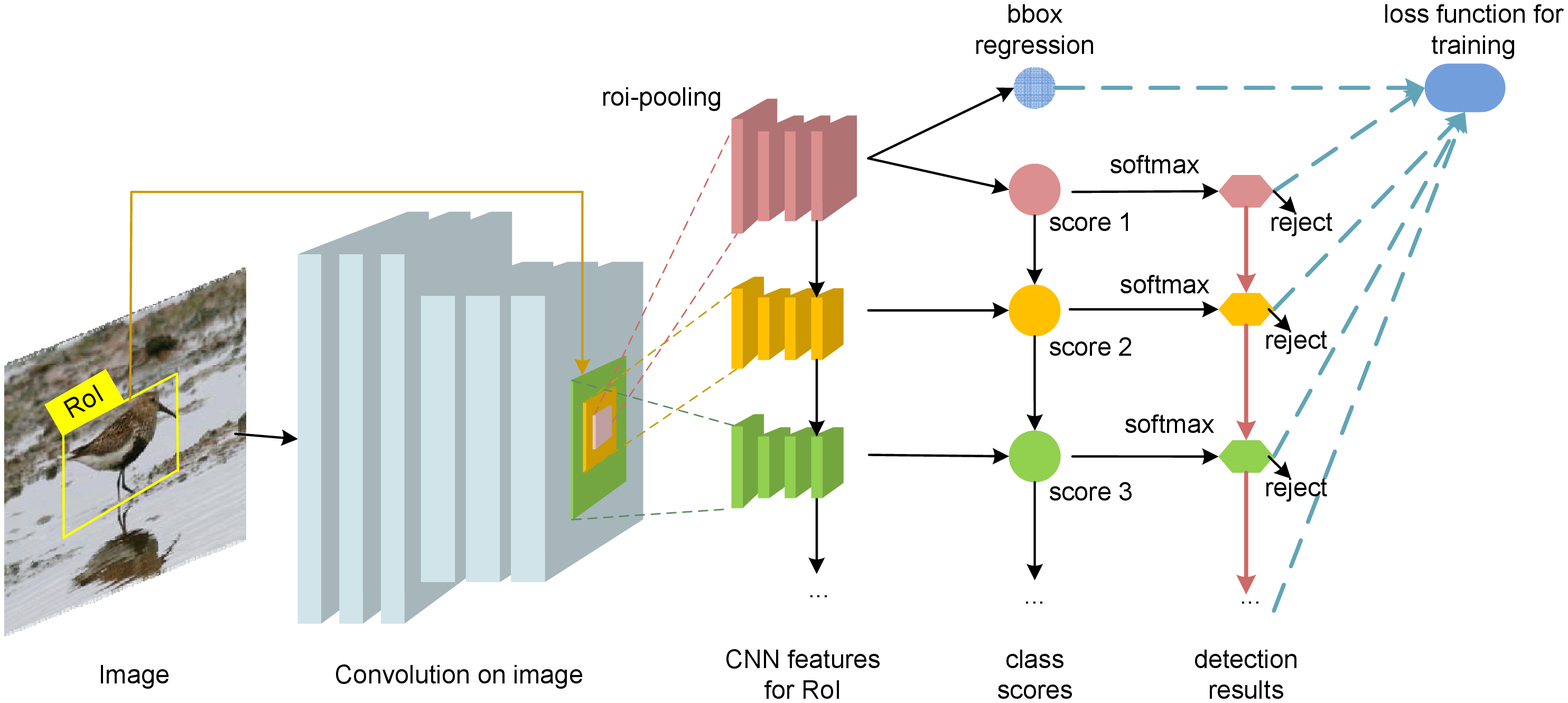,width=18cm}}
\end{center}
\vspace{-10pt}
   \caption{Overview of the CC-Net. Several convolutional layers are used on the input image, then roi-pooling is used for obtaining features of different resolutions and contextual regions. These features are passed to several convolutional layers. Then features in different stages are integrated by feature chaining and classification chaining for obtaining the detection results. At the training and testing stage, easy background samples are rejected at early stages. Bounding box regression and cascaded classifiers are jointly learned. Best viewed in color.}
\label{fig:overview}
\end{figure*}

\begin{figure*}
\begin{center}
\centerline{\epsfig{figure=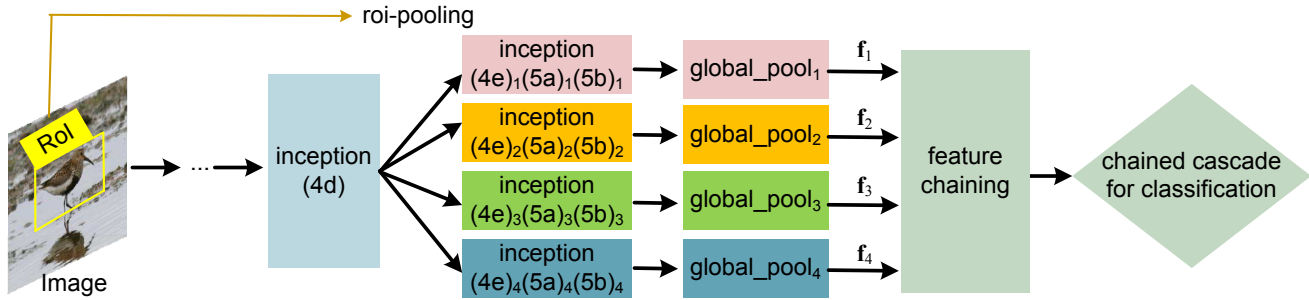,width=18cm}}
\end{center}
\vspace{-10pt}
   \caption{An example of the CC-Net based on the BN-Inception net.  Best viewed in color.}
\label{fig:SC_example}
\end{figure*}

\section{The CC-Net for object detection}

\subsection{Brief introduction of the fast RCNN}
This paper adopts the fast RCNN framework for object detection. In this approach, 1) a set of regions of interest (RoIs) are generated by a region proposal approach; 2)  CNN feature maps for the input image are generated by several convolutional layers;  3) the roi-pooling layer projects the RoIs onto the CNN feature maps and extracts feature maps of the same size for RoIs of different sizes; 4) the layers after roi-pooling are conducted to obtain the final features, from which the classification scores and the regressed coordinates for bounding-box relocalization are predicted. 
%The existing CNN based approaches in \cite{yang2016craft, gidaris2016attend} are used for generating RoIs in our implementation. 
%With the RoIs, the proposed cascaded CNN is used for predicting the classification score and the regressed coordinates for bounding-box re-localization.

\subsection{Overview of the CC-Net}
Fig. \ref{fig:overview} shows an overview of the chained cascade ConvNet (CC-Net). The existing CNN based approaches in \cite{yang2016craft, gidaris2016attend} are used for generating RoIs in our implementation. 
 Based on the fast RCNN framework, it uses several convolutional layers for extracting convolutional features from the image and then use roi-pooling for projecting features of RoIs into the same size. These features are then used by the chained features and classifiers with multiple stages for classification. In our implementation, the BN-Inception in \cite{ioffe2015batch} is used as the baseline network if not specified. If only single stage of features and classifiers is used, then Fig. \ref{fig:overview}  becomes a fast RCNN implementation of the BN-Inception model. The major modifications to fast RCNN are as follows:
\begin{itemize}
\item Chained cascade of classifiers with several stages are used for object detection. At each stage, easy background samples are rejected. Samples not rejected go to the next stage. By classifier chaining, the classification scores in the previous stages are used for  classification in the current stage.
\item Classifiers at different stages use different features. These features can be the same in CNN structure but different in learned parameters, resolution and contextual regions.
\item The features in previous stages are chained with the CNN features at the current stage. With this chaining, the features at previous stages serve as the prior knowledge for the features at the current stage. 
\item The bounding box regressor, chained classifiers and features are learned end-to-end through back-propagation from a single loss function.
\end{itemize}

Fig. \ref{fig:SC_example} shows an example of the CC-Net based on the BN-Inception Net. There are nine inception modules in the BN-Inception Net, the roi-pooling layer is placed after the sixth module, which is inception (4d). Roi-pooling is used for obtaining features of different resolutions and contextual regions. The features after roi-pooling for stage $t$ is denoted by $\mathbf{h}_t$, $t=1, 2, 3, 4$. At stage $t$, the features in $\mathbf{h}_t$ go through the remaining inception modules and global average pooling for obtaining features $\mathbf{f}_t$. Then these features are combined by feature chaining, with details in Section \ref{Sec:FeatChain}. The chained features are then used by chained cascade of classification for detecting objects, with details in Section \ref{Sec:ClsChain}.

\subsection{Feature chaining}
\label{Sec:FeatChain}
\subsubsection{Preparation of features with diversity}
\begin{figure}
\begin{center}
\centerline{\epsfig{figure=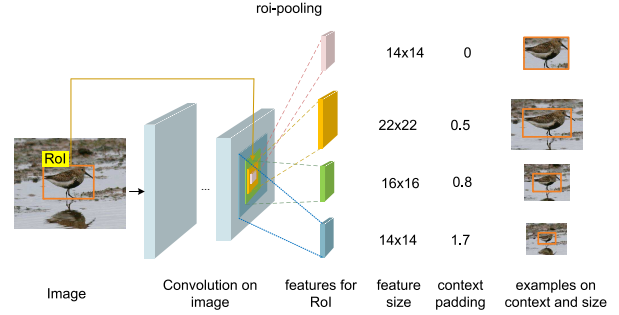,width=8cm}}
\end{center}
\vspace{-10pt}
   \caption{The use of roi-pooling to obtain features with different resolutions and contextual regions. After roi-pooling features in different stages have different sizes and contextual padding value $c$. The original box size is used when $c=0$. And 1.5 times the original box size is used when $c=0.5$. Best viewed in color.}
\label{fig:Feature}
\end{figure}

Cascaded classifiers in different stages can use different features. Multi-region, multi-context features were found to be effective in \cite{bell2015inside, gidaris2015object, zeng2016gated}. In order to obtain features with diversity, we apply roi-pooling from image features using different contextual regions and resolutions.  For a standard input image with size $224\times 224$, the Inception (4d) outputs feature maps of size $14\times 14$. Therefore, the features after roi-pooling should have size $14\times 14$ if the fast RCNN approach is adopted. In our CC-Net, however, the roi-pooled features have the same number of channels but have different sizes at different stages. The sizes of roi-pooled features are respectively $14\times 14$, $22\times 22$, $16\times 16$ and $14\times 14$ for features at stages 1, 2, 3 and 4. The contextual regions for these features are also different. Suppose the RoI has width $W$ and height $H$.  Denote $c$ as the context padding value for the RoI. The padded region has the same center as the RoI and has width $(1+c)\cdot W$ and height $(1+c)\cdot H$. $c=0, 0.5,  0.8$, and 1.7 for stages 1, 2, 3, and 4 respectively.  Figure \ref{fig:Feature} shows the contextual regions for features at different stages.  These features are arranged with increasing contextual regions.

After features of different resolutions and contextual regions are obtained, they go through the remaining three inception modules (4e), (5a) and (5b). In order to increase the variation of features, the inception modules  at different cascade stages have different parameters. Denote the inception module (4e) at stage $t$ by  (4e)$_t$.  The modules (4e)$_1$, (4e)$_2$, (4e)$_3$, and (4e)$_4$  are initialized from the same pretrained inception module (4e) but have different parameters during the finetuning stage because they receive different gradients in  backpropagation. The treatment for the module (4e)$_t$ are also applied for the inception modules (5a)$_t$ and (5b)$_t$. The CNN features obtained from inception modules (5b)$_t$ have different sizes. We use global average pooling for these features so that they have the same size before feature chaining.

%Although the descriptions in this section is specific for the BN-Inception net. This design in obtaining features with different resolutions, sizes and contextual regions and using different parameters in different stages can be easily applied for other networks. 

\subsubsection{The feature chaining structure}
\begin{figure}
\begin{center}
 \centerline{\epsfig{figure=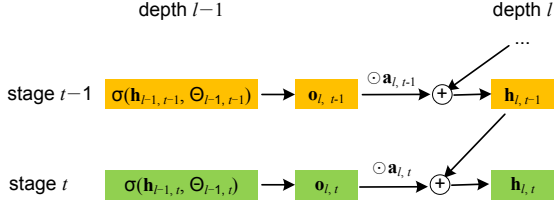,width=8cm}}
\end{center}
\vspace{-10pt}
   \caption{Feature chaining. Features $\mathbf{h}_{l,t}$ with depth $l$ and stage $t$ are obtained by summing up the features in the previous stage $\mathbf{h}_{l,t-1}$ and nonlinear mapping results from the features in the previous layer  $\mathbf{h}_{l-1,t}$.  }
\label{fig:Cascade}
\end{figure}

Denote the features at depth $l$ and stage $t$ as $\mathbf{h}_{l,t}$.
In order to use the features in previous stages as the prior knowledge when learning features for stage $t$, we design the feature chaining which has the following formulation:
\begin{align}
\mathbf{h}_{l,t} =\mathbf{h}_{l,t-1} +  \mathbf{a}_{l,t} \odot \mathbf{o}_{l,t},  \label{eq:cascadefeat}\\
 \mathbf{o}_{l, t} = \sigma(\mathbf{h}_{l-1,t}, \Theta_{l-1,t}),
 \label{eq:cascadefeat2}
\end{align}
where $\mathbf{a}_{l,t-1}$ and $ \Theta_{l-1,t}$ are parameters learned from the network. In this design, the $\mathbf{h}_{l,t} $ is obtained from the features in previous stages $\mathbf{h}_{l,t-1}$ and nonlinear mapping of the features from the shallower layer $\mathbf{h}_{l-1,t}$. $\mathbf{a}_{l,t-1}$ denotes a vector of scalers for scaling the features $ \mathbf{h}_{l,t-1}$ in the previous stage. The operation $\odot$ in (\ref{eq:cascadefeat}) denotes dot product, where $[\alpha_1 \  \alpha_2]\odot [\beta_1 \ \beta_2] = [\alpha_1\beta_1 \ \alpha_2 \beta_2]$.  The elements in $\mathbf{a}_{l,t-1}$ are initialized as 1 and are learned through backpropagation to control the scale of the features. 
The nonlinear mapping in $\sigma(\mathbf{h}_{l-1,t}, \Theta_{l-1,t})$ in (\ref{eq:cascadefeat2}) can be implemented by convolutional layer or fully connected layer with nonlinear activation function. Fig. \ref{fig:Cascade} (a) shows the block diagram for this structure. 

\begin{figure}
\begin{center}
 \centerline{\epsfig{figure=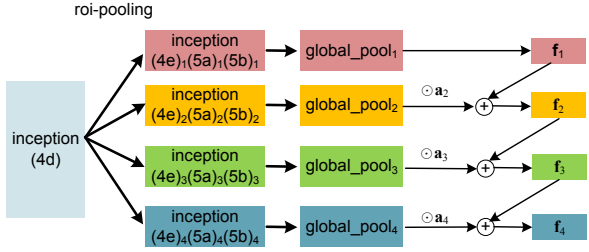,width=8cm}}
\end{center}
\vspace{-10pt}
   \caption{Chaining features for BN-inception. }
\label{fig:CascadeBN}
\end{figure}

In our implementation based on the BN-Inception net as shown in Fig. \ref{fig:CascadeBN}, feature chaining is placed after the global average pooling, where all features are spatially pooled to have spatial size $1\times 1$  and 1024 channels. 
 Denote the features after global pooling  for stage $t$ as $\mathbf{o}_t$.
 The following procedure is used for obtaining the chained features:
\begin{equation}
\begin{split}
\mathbf{f}_{1} &=  \mathbf{o}_1, \\
\mathbf{f}_{2} &=  \mathbf{o}_{2} \odot  \mathbf{a}_2 + \mathbf{f}_1, \\
\mathbf{f}_{3} &=  \mathbf{o}_{3} \odot  \mathbf{a}_3 + \mathbf{f}_2, \\
\mathbf{f}_{4} &=  \mathbf{o}_{4} \odot  \mathbf{a}_4 + \mathbf{f}_3.\\
\end{split}
\end{equation}
In this implementation, the feature $\mathbf{f}_{t}$ at stage $t$ is obtained by summing up features $\mathbf{f}_{t-1}$ in the previous stage and the $\mathbf{a}_2 \odot \mathbf{o}_2$, which is the output from the previous layer global\_poo$_t$ weighted by $\mathbf{a}_2$.
The summed features $\mathbf{f}_t, t= 1, 2, 3, 4$ are then used for chained cascade of classification.

\subsubsection{Discussion}
Feature chaining includes the concept of stage.  Features $\mathbf{h}_{l, t}$ and $\mathbf{h}_{l, t+1}$ have the same depth but are different in stages. Features in different stages have specific objectives -- they are used by classifiers for rejecting easy background samples. The features of the same depth but different stages communicate through feature chaining.

\begin{figure}
\begin{center}
 \centerline{\epsfig{figure=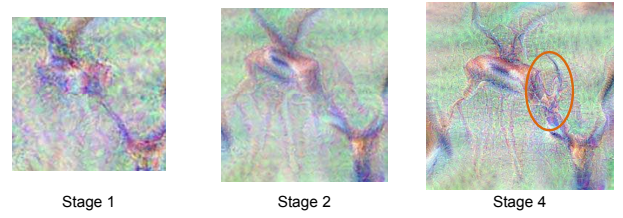,width=8cm}}
\end{center}
\vspace{-10pt}
   \caption{Visualization of the object class antelope for the learned model at the stages 1, 2, and 4 using the DeepDraw \cite{web:deepDraw}. }
\label{fig:Visualize}
\end{figure}

With feature chaining, features at the current stage take the features in previous stages into consideration. Therefore, the CNN layers at the stage $t$ no longer need to represent the information existing in previous stages. Instead, they will focus  on representations that are complementary to those in previous stages. Fig. \ref{fig:Visualize} shows the visualization of the learned model for the object class antelope in different cascade stages. At stage 1, the learned feature for antleope is very rough. The learned feature at stage 4 looks into more details when compared with the feature at stage 2. There are many possible locations of head and horns in the visualized result. The shape for head and horn highlighted by the ellipse for the antelope at stage 4 does not exist in stage 2. 

%Feature chaining is different from simple increase of depth or width. 

%Feature chaining is an efficient way of passing messages among features in different cascade stages.  
%In existing approaches like recurrent neural networks (RNN), when a message is passed from a layer with $C_1$ channels to another layer with $C_2$ channels, the number of parameters is proportional to $C_1C_2$. In feature chaining, however, only parameters $\mathbf{a}_{l,t}$ are used, the number of parameters of which is proportional to $C_1$. Therefore, feature cascade saves the parameter size in message passing by $C_2$ times, which is normally hundreds or thousands. For example, $C_1=C_2=1024$ in our implementation for BN-Inception model.

%We freeze the parameters in all Batch Normalization (BN) layers at the finetuning stage because this freezing is found to achieve better results. If BN layer is not frozen in other applications, the scaling vector $\mathbf{a}_{l,t}$ learned in feature chaining can be merged into the scale learning in BN layer. In this case, the scale learning and its corresponding parameters are no longer needed. If BN layer is frozen and the elements in scaling vector  $\mathbf{a}_{l,t}$ is fixed as 1 for finetuning, there will be 4\% mAP drop in detection accuracy. Therefore, empirical results show the need to learn the scaling of features.

\subsection{Cascade chaining for classification in CC-Net}
\label{Sec:ClsChain}
\subsubsection{Cascade chaining for classification}
This section briefly introduces cascade chaining for  binary classifiers, which is called soft cascade in \cite{bourdev2005robust} and boosting chain in \cite{xiao2003boosting}. 
Denote $\mathbf{f}_t$ as the features for the classifier at stage $t$, $t=1, 2, \ldots, T$. Denote $c_t(\mathbf{f}_t)$ as the classification function for the feature $\mathbf{f}_t$ at the stage $t$. 
%Boosting was used for learning this classfier in \cite{bourdev2005robust, xiao2003boosting} for face detection.
The partial sum of classification scores up to and including the $t$th stage is defined as follows:
\begin{equation}
ps_t = \sum_{i=1, \ldots, t} c_t(\mathbf{f}_t).
\end{equation}
In cascade chaining, the partial sum $ps_t$ is compared with the threshold $r_t$. If  $ps_t<r$, then the sample is not considered as an object. Otherwise, the next stage of comparison is performed. If the sample is not rejected after $T$ stages of such rejection scheme, the score $ps_T$ will be used as the detection score. This approach is summarized in Algorithm \ref{alg:cascade}.
The main difference between cascade chaining and conventional cascade is that conventional cascade only uses $c_t(\mathbf{f}_t)$ as the score at the stage $t$ but cascade chaining includes the previous scores. If the $ps \leftarrow  ps +  c_t(\mathbf{f}_t)$ in Algorithm \ref{alg:cascade}  is replaced by $ps \leftarrow  c_t(\mathbf{f}_t)$, then cascade chaining degenerates to conventional cascade.

%The cascade chaining for one class of object can be easily extended to multiple classes by treating each class of object separately.

\begin{algorithm}[b]
\KwIn{$ \Psi = \{ \mathbf{f}_t \}$, features at stage $t$ for a given sample.\\
}
\KwOut{$p$, the predicted detection score for the sample.}
$ps \leftarrow 0$\\
\textbf{for} $t=1 \ldots T$ \\
 ~~~~~$ps \leftarrow  ps +  c_t(\mathbf{f}_t)$ . \\
 ~~~~~ If $ps<r_t$,  return $-\infty$.  \\
\textbf{end for}\\
return $ps$
  \caption{The cascade chaining algorithm using classification function $c_t(*)$  and threshold $r_t$ at stage $t$ for rejecting samples at early stages.}
\label{alg:cascade}
\end{algorithm}

\subsubsection{Cascade chaining at the testing stage in CC-Net}
\begin{figure}
\begin{center}
\centerline{\epsfig{figure=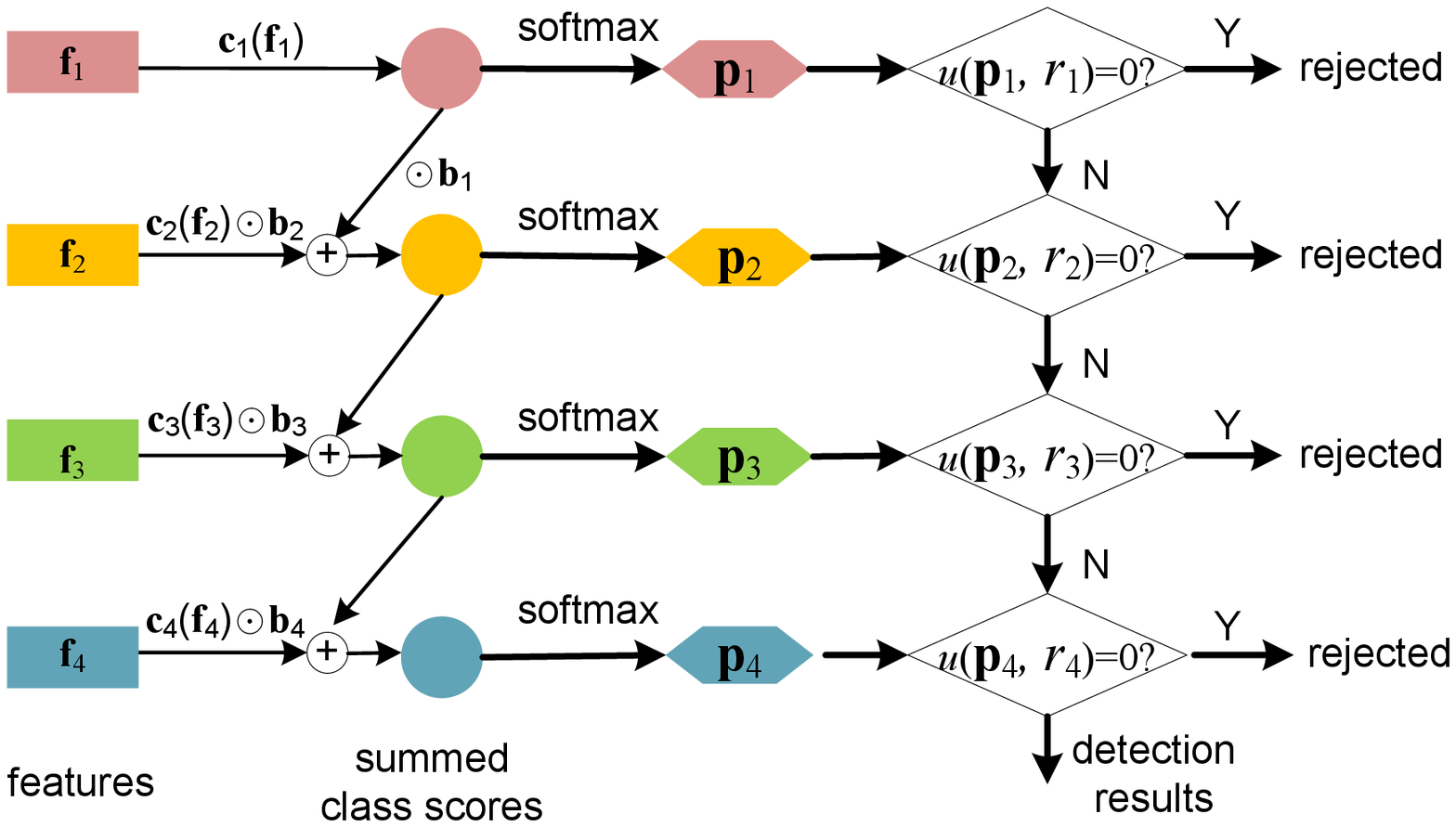,width=8cm}}
\end{center}
\vspace{-10pt}
   \caption{The cascade chaining at the testing stage. Given features $\mathbf{f}_t$, an fc layer is used for obtaining classification score at stage $t$. The classification sores from previous stages are combined with the scores at the current stage to obtained the summed scores. The summed scores undergo softmax to obtain the normalized scores $\mathbf{p}_t$ at stage $t$. Then the thresholding function $u(\mathbf{p}_t, r_t)$ decides whether to reject the sample or not. The sample not rejected after $T$ stages uses the $\mathbf{p}_T$ as the detection result. $T=4$ in the figure. Best viewed in color.}
\label{fig:testSC}
\end{figure}

In the CC-Net, the partial sum of classification scores up to and including the $t$th stage is obtained from the set of features $\{\mathbf{f}_t\}$ as follows:
\begin{align}
\tilde{\mathbf{p}}_t &= [p_{t, 1}\  \ldots \ p_{t, K+1}] =  \left( \sum_{i=1, \ldots, t} \mathbf{b}_t \odot \mathbf{c}_t(\mathbf{f}_t) \right).  \label{eq:cascade1} 
\end{align}
The $\mathbf{c}_t(\mathbf{f}_t)$ in (\ref{eq:cascade1}) denotes the $K+1$-class classifier which takes the feature $\mathbf{f}_t$ as input and outputs $K+1$ classification scores on the input sample being one of the $K$ classes or background.  $\mathbf{c}_t(\mathbf{f}_t)$ is implemented using the fully connected (fc) layer in the CC-Net. The $\sum$ in (\ref{eq:cascade1}) denotes the summation over vectors. The operation $\odot$ in (\ref{eq:cascade1}) denotes dot product, where $[\alpha_1 \  \alpha_2]\odot [\beta_1 \ \beta_2] = [\alpha_1\beta_1 \ \alpha_2 \beta_2]$. $\mathbf{b}_t$ for this dot product is the vector of scaling parameters for controlling the scale of the classification scores. The scores $\tilde{\mathbf{p}}_t$ in (\ref{eq:cascade1} ) are normalized to probabilities $\mathbf{p}_t $ using the softmax function as follows:
\begin{align}
\mathbf{p}_t &=  [{p}_{t, 1}\  \ldots \ {p}_{t, K+1}] =\textrm{softmax} (\tilde{\mathbf{p}}_t), \label{eq:cascade2}  \\
\textrm{where } {p}_{t, k} &= {\tilde{p}_{t, k}}/{ \sum_{k=1}^{K+1}\tilde{p}_{t, k}}. \label{eq:cascade3}
\end{align}
The probabilities $\mathbf{p}_t$ are used by the following thresholding function for deciding whether to reject the given sample or not:
\begin{align}
u(\mathbf{p}_t, r_t) =
\begin{cases} 
1,& \quad \textrm{if }  \max\{ p_{t, 1}\  \ldots \ p_{t, K} \} > r_t, \\
0,& \quad \textrm{ otherwise.} \\
\end{cases}
  \label{eq:cascade4}
\end{align}
If $u(\mathbf{p}_t, r_t) =0$, then the sample is considered as a background and rejected. The classifiers at latter stages are not used for saving testing time. If the sample is not rejected after $T$ iterations, then $\mathbf{p}_T$ is used as the detection result. Fig. \ref{fig:testSC} shows the diagram for cascade chaining at the testing stage in  CC-Net. %$T=4$ in our implementation.

\begin{table*}[t]
\centering
{\small
\begin{tabular}{cccccccccc}
\hline
appraoch& RCNN & Berkeley & GoogleNet &  DeepID-Net&Superpixel &ResNet&GBD-Net & CC-Net \\
&\cite{girshick2014rich}&\cite{girshick2014rich}&\cite{szegedy2015going}&\cite{ouyang2015deepid}&\cite{yan2015object}&\cite{he2016deep}&\cite{zeng2016gated}&\\
\hline
val2(sgl)&31.0&33.4&38. 5&48.2&42.8&60.5 &51.4&54.5\\
\hline
\end{tabular}
\caption{Object detection mAP (\%) on ImageNet val2 for state-of-the-art approaches with single model.}
\label{table:overall}
}
\end{table*}

\begin{table*}[]
\centering
{\small
\setlength{\tabcolsep}{1pt}
\begin{tabular}{c|c|c|c|cccccccccccccccccccc}
\hline
method                                & network & region                                & mAP  & aero & bike & bird & boat & bottle & bus  & car  & cat  & chair & cow  & table & dog  & horse & mbike & persn & plant & sheep & sofa & train & tv   \\
 \hline
FRCN                                  & VGG16   & SS  & 70.0   & 77.0   & 78.1 & 69.3 & 59.4 & 38.3   & 81.6 & 78.6 & 86.7 & 42.8  & 78.8 & 68.9  & 84.7 & 82.0    & 76.6  & 69.9  & 31.8  & 70.1  & 74.8 & 80.4  & 70.4 \\
MR      & VGG16   & SS  & 78.2 & 80.3 & 84.1 & 78.5 & 70.8 & 68.5   & 88.0   & 85.9 & 87.8 & 60.3  & 85.2 & 73.7  & 87.2 & 86.5  & 85.0    & 76.4  & 48.5  & 76.3  & 75.5 & 85.0    & 81.0   \\
OHEM  & VGG16   & SS  & 78.9 & 80.6 & 85.7 & 79.8 & 69.9 & 60.8   & 88.3 & 87.9 & 89.6 & 59.7  & 85.1 & 76.5  & 87.1 & 87.3  & 82.4  & 78.8  & 53.7  & 80.5  & 78.7 & 84.5  & 80.7 \\
FRCN                                  & BN      & AT          & 77.6 & 81.0   & 80.3 & 79.4 & 71.6 & 65.5   & 87.2 & 86.4 & 89.9 & 58.8  & 82.8 & 69.6  & 88.1 & 87.7  & 78.7  & 79.9  & 48.9  & 78.5  & 74.1 & 85.3  & 78.4 \\
ours                                  & CC-Net  & AT          & 81.1 & 80.9 & 84.8 & 83   & 75.9 & 72.3   & 88.9 & 88.4 & 90.3 & 66.2  & 87.6 & 74.0    & 89.5 & 89.3  & 83.6  & 79.6  & 55.2  & 83.4  & 81.0   & 87.8  & 80.7 \\
 \hline
\end{tabular}
%\begin{tabular}{cccc|c}
%method     & Train set & Network                  & Region proposal & mAP  \\
%\hline
%FRCN & 07+12     & VGG16  \cite{simonyan2014very}                  & SS \cite{Smeulders:SelectiveSearch}              & 70   \\
%FRCN & 07+12     & BN  \cite{ioffe2015batch}           & AttractioNet \cite{gidaris2016attend}   & 78.0 \\
%Ours & 07+12     & BN + CC-Net & AttractioNet\cite{gidaris2016attend}    & 81.1 \\
%\end{tabular}
\caption{Object detection mAP (\%) on VOC 2007 for fast RCNN (FRCN) with different settings. \emph{method} denotes using the method FRCN \cite{girshick2015fast}, MR \cite{gidaris2015object}, OHEM \cite{shrivastava2016training}, or ours.  Multi-scale and iterative bbox regression are used for MR and OHEM but not  for the other methods. \emph{network} denotes the use of the 16-layer VGG model (VGG16) \cite{simonyan2014very}, the BN-inception model (BN) \cite{ioffe2015batch}, or the baseline BN-inception with our design of CC-Net (CC-Net). \emph{region} denotes the use of selective search (SS) \cite{Smeulders:SelectiveSearch}  or AttractioNet (AN) \cite{gidaris2016attend}  for region proposal. }
\label{Tab:VOC07}
}
\vspace{-10pt}
\end{table*}

\subsubsection{Training CC-Net}
A multi-task loss of classification and bounding-box regression is used to jointly optimize the CC-Net. Suppose there are $K$ object classes to be detected. Denote the set of estimated class probabilities for a sample  by $\mathbf{p}=\{\mathbf{p}_t| t= 1, \ldots, T\}$, where $\mathbf{p}_t=[p_{t, 0}\ \ldots p_{t, K} ]$ is the estimated probability vector at stage $t$ and $p_{t, k}$ is the estimated probability for the $k$th class. $k=0$ denotes the background. $\mathbf{p}_t$ is obtained by a softmax over the $K+1$ outputs of a fc layer. Another layer outputs bounding-box regression offsets $\mathbf{l} =\{\mathbf{l}^{k}|k=1, \ldots $K$ \}$, $\mathbf{l}^k=(l^k_\textrm{x}, l^k_\textrm{y}, l^k_\textrm{w}, l^k_\textrm{h})$  for each of the $K$ object classes, indexed by k. Parameterization for $\mathbf{l}^k$ is the same as that in \cite{girshick2014rich}.
The loss function is defined as follows:
\begin{align}
&L(\mathbf{p}, k^*, \mathbf{l} , \mathbf{l} ^*) = L_{cls}(\mathbf{p}, k^*)+L_{loc}(\mathbf{l}, \mathbf{l}^*, k^*), \\
& L_{cls}(\mathbf{p}, k^*)=-\sum_{t=1}^T \lambda_t u_t \log p_{t,k^*}, \\
&u_t = \prod_{i=1}^{t-1} [p_{i,k^*}<r_i] \ \textrm{when } t>1,  \ \ \ \ u_1 = 1.
 \label{eq:loss}
\end{align}
$L_{cls}(*)$ is the loss for classification and $L_{loc}$ is the loss for bounding-box regression. If $\lambda_t = u_t=1$ and $T=1$, then $L_{cls}(*)$  is a normal cross entropy loss.
$u_t$ evaluates whether the sample is rejected in the previous stages. If a sample is rejected in the previous stage, it is no longer used for learning the classifier in the current stage. Since we did not constrain the sample to be background for rejection, easy positive samples are also rejected at early stages during training. $\lambda_t$ is a hyper parameter that controls the weight of loss for each stage of cascaded classifier. We set $\lambda_T=1$ and $\lambda_t=0.02/T$ for $t=1, \ldots T-1$. Loss is used for $t=1, \ldots T-1$ so that the learned classifiers in these stages can learn reasonable classification scores for rejecting background samples.  Since the score in the last classifier is used as the final detection score, the classification loss in the last stage has much higher weight than the loss in other stages. For $L_{loc}$, we use the  smoothed $L_1$ loss in \cite{girshick2015fast}. With this loss function, bounding box regression, chained features and all cascaded classifiers are learned jointly through backpropagation.

\section{Experimental results}
\subsection{Experimental setup}
The CC-Net is implemented based on the fast RCNN pipeline. The BN-Inception net is used as the baseline network if not specified.
In the CC-Net, the feature chaining is used after the global average pooling of the BN-Inception net \cite{ioffe2015batch} as shown in Fig. \ref{fig:CascadeBN}. In the CC-Net, the layers belonging to the baseline networks are initialized by these baseline networks pre-trained on the ImageNet dataset. The parameters $\mathbf{a}_t$ in feature chaining and classification chaining are initialized as 1. 
 For region proposal, we use the Craft in \cite{yang2016craft} for ImageNet and the AttractionNet in \cite{gidaris2016attend} for VOC 2007 if not specified. 

We evaluate our method on two public object detection datasets, ImageNet \cite{ILSVRC15} and PASCAL VOC 2007 \cite{Everingham:PacalVOC}. Since the ImageNet object detection task contains a sufficiently large number of images and object categories to reach a conclusion, evaluations on component analysis of our training method are conducted on this dataset. This dataset has 200 object categories and consists of three subsets. i.e., train, validation and test data. We follow the same setting in \cite{girshick2014rich} and split the whole validation subset into two subsets, val1 and val2.  The network finetuning step uses training samples from train and val1 subsets. The val2 subset is used for evaluating components. For all networks, the learning rate and weight decay are fixed to 0.0005 during training. For the VOC07 dataset, we train on VOC07+12 training data and test on the VOC07 testing data. All our results are for single model with single-scale training and testing. Single-stage bounding box regression is used.

\subsection{ImageNet results}
On this dataset, we compare with the top methods tested on the val2 dataset. We compare our framework with several other state-of-art approaches \cite{girshick2014rich,szegedy2015going,ioffe2015batch,ouyang2015deepid,yan2015object,he2016deep, zeng2016gated}. The mean average precision for these approaches are shown in Table \ref{table:overall}.  Our work is trained using the provided data of ImageNet.  Compared with these approaches, our single model result ranks No. 2, lower than the ResNet \cite{he2016deep} which uses a much deeper network structure. 

\subsection{PASCAL VOC 2007 results}
On this dataset, the VOC07+12 trainval dataset are used for training and the VOC07 test set is used for evaluation. As shown in Table \ref{Tab:VOC07}, the baseline BN-inception model has mAP 77.6\% when AttractioNet is used for region proposal \cite{gidaris2016attend}. With our design in chaining features and cascaded classifiers, the mAP is 81.1\%. We also list some of the recent approaches  using multi-region features \cite{gidaris2015object} and hard negative mining \cite{shrivastava2016training} for comparison.

\subsection{Component analysis}
\subsubsection{Baseline BN-Inception with different region proposals}
The experimental results for the baseline BN-Inception using different region proposals are summarized in Table \ref{Table:RPN}. It is reported in \cite{zeng2016gated} that the BN-Inception with Craft \cite{yang2016craft}  for region proposal has mAP 46.3\% on ImageNet. The authors in \cite{yang2016craft} have provided online better region proposal results for ImageNet, for which the baseline BN-Inception we implemented has mAP 49.4\%. Similarly, we choose better region proposal for VOC07. The baseline BN-Net has mAP 73.1\% when combining the selective search \cite{Smeulders:SelectiveSearch} and Edgebox \cite{ZitnickDollarECCV14edgeBoxes} for region proposal  and has mAP 77.6\% when using the AttractionNet \cite{gidaris2016attend} for region proposal. Since the region proposals from Craft+ and AttractionNet are shown to be effective, we have used Craft+ for ImageNet and AttractionNet for VOC07 as the better baseline for all our results.

\begin{table}[]
\centering
{\small
\begin{tabular}{c|cc}
\hline
Region proposal & Craft \cite{yang2016craft}                                                        & Craft+ \cite{yang2016craft}            \\
mAP on ImageNet  & 46.3                                                                                & 49.4                                     \\
\hline
Region proposal & SS \cite{Smeulders:SelectiveSearch}+Edgebox \cite{ZitnickDollarECCV14edgeBoxes} & AttractionNet \cite{gidaris2016attend} \\
mAP on  VOC07 & 73.1            & 77.6 \\
\hline
\end{tabular}
\caption{Baseline BN-Inception model with different region proposals. Craft+ is the new region proposal based on Craft provided by the authors.}
\label{Table:RPN}
}
\end{table}

\subsubsection{Results on cascade chaining} 
In order to evaluate the performance gain from chaining cascaded classifiers, we use the BN-Inception as the baseline. Multi-context multi-resolution features are not included. For the results in Table \ref{Tab:Cascade}, all cascaded classifiers take the output the global\_pool layer in BN-Inception as the feature. Features and classifiers are jointly learned. Compared with the baseline,  online hard example mining (OHEM) \cite{shrivastava2016training} improves mAP by 0.8\%, adding two extra stages of cascaded classifiers improves the mAP by 1.1\%, and adding four extra cascaded classifiers improves mAP by 1.5\%. The use of more cascaded classifiers provides better detection accuracy.  If the 4 extra stages of cascade do not use chaining, \ie not using previous classification scores for the current classification score, there will be 0.4\% mAP drop.
In this experiment, we use the region proposal of Craft+. Only 300 boxes per image are used for both training and testing. This experiment shows that OHEM and cascade provide improvement. The improvement is not so large as that in \cite{shrivastava2016training}, possibly because 2000 boxes per image from selective search \cite{Smeulders:SelectiveSearch} were used in \cite{shrivastava2016training}, which contains more background samples that should be rejected. Increasing the number of boxes from Craft+ does not result in mAP gain, because Craft+ has sufficiently high recall on these 300 boxes.

\begin{table}[]
\centering
\begin{tabular}{c|ccccc}
\hline
+ 2 cascade stages?  &      &            & \checkmark &            &            \\
+ 4 cascade stages?  &      &            &            & \checkmark & \checkmark \\
OHEM \cite{shrivastava2016training}?                &      & \checkmark &            &            &            \\
cascade?             &      &            & \checkmark & \checkmark & \checkmark \\
chaining classifier? &      &            & \checkmark & \checkmark &            \\
\hline
  mAP                   & 49.4 & 50.2       & 50.5       & 50.9       & 50.5   \\
                     \hline
\end{tabular}
\caption{ImageNet val2 detection mean average precision (\%) for baseline BN-Inception with different setup on cascade or online hard example mining (OHEM). `chaining classifier' denotes the result using the chaining for classifier, in which scores in previous stages are used for the current stage. `cascade' denotes the use of cascade.}
\label{Tab:Cascade}
\end{table}

%\begin{figure}
%\begin{center}
%   \includegraphics[width=1\linewidth]{./Accuracy_Complextity.pdf}
%\end{center}
%\vspace{-10pt}
%   \caption{The number of remaining object boxes after cascade in three stages. By using chained cascade with multiple stages, it is possible to conduct trade-off between accuracy and speed. }
%\label{fig:Accuracy_Complextity}
%\end{figure}

\subsubsection{Chaining features and classifiers}
Table \ref{Table:Cascade2} shows the performance for different settings in chaining features and classifiers.
Multi-region features are found to  be effective in \cite{gidaris2015object}. When we concatenate features of different contextual regions and resolutions but without the feature chaining or the classification cascade,  the mAP is 50.5\%. It uses the same features as the CC-Net. Based on these features, the use of cascade chaining improves the mAP by 0.8\%. Based on these features, mAP is 54.5\% if both feature chaining and cascade chaining are used in the CC-Net.  Based on the diverse features with cascade chaining, the inclusion of feature chaining in the CC-Net improves the mAP by 3.2\%. Removing the chaining of classifiers from the CC-Net results in 0.8\% mAP drop.

When learning the chaining of features and classifiers, scaling vectors $\mathbf{a}$ and $\mathbf{b}$ are used for controling the scales of features and classification scores, if these scalers are fixed as 1 but not learned, the mAP will drop by 1.7\%. No mAP gain is observed when the scaling vector $\mathbf{a}$ for feature chaining with $C_1$ parameters is replaced by fully connected layer with $C_1 C_2$ parameters.

\begin{figure}
\begin{center}
 \centerline{\epsfig{figure=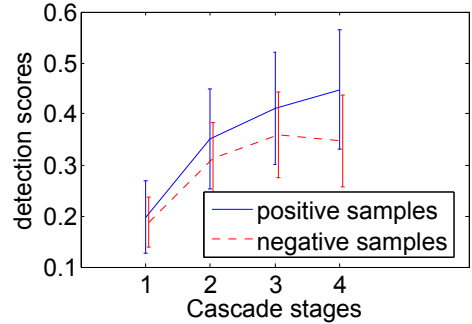,width=6.5cm}}
\end{center}
\vspace{-10pt}
   \caption{Average scores and variances of positives samples and negative samples. By using soft cascade with multiple stages, it is easier to distinguish object samples from background samples.  Best viewed in color.}
\label{fig:Cascade_scores}
\end{figure}

Fig. \ref{fig:Cascade_scores} shows the average of scores for the first 10,000 object samples and background samples in ImageNet val2. As shown in Fig. \ref{fig:Cascade_scores}, with the cascade of classifiers in more stages, the detection scores are better in distinguishing positive samples and negative samples. 

%The cascade allows for trade-off between computational complexity and accuracy. With 0.6\% mAP drop, the number of remaining  boxes after the cascade in the first stage can be reduced to around 50\%.

\begin{table}[]
\centering
\begin{tabular}{c|ccccc}
\hline
diverse features?     &      & \checkmark & \checkmark & \checkmark & \checkmark \\
cascade?          &      &            & \checkmark & \checkmark & \checkmark \\
classifier chaining? &      &            & \checkmark & \checkmark &            \\
features chaining?    &      &            &            & \checkmark & \checkmark \\
\hline       mAP               & 49.4 & 50.5       & 51.3       & 54.5       & 53.7     \\
\hline
\end{tabular}
\caption{ImageNet val2 detection mean average precision (\%) for baseline BN-Inception with different setting on feature chaining and classifier chaining. `cascade' denotes the use of cascade.}
\label{Table:Cascade2}
\end{table}

\section{Conclusion}
In this paper, we present a chained cascade neural network (CC-Net) for object detection. In this net, the cascade of classifiers in multiple stages are jointly learned through a single end-to-end neural network. This network includes classifier chaining, in which classifier at the current stage takes the classification scores in previous as prior knowledge. We further propose feature chaining, which uses the features in previous stages as the prior information for the features in the current stage. The effectiveness of CC-Net is validated on ImageNet and VOC 2007 object detection datasets.

{\small
\bibliographystyle{ieee}
\bibliography{./PME}
}

\end{document}